\documentclass[letterpaper]{article} 
\usepackage{aaai2027} 
\nocopyright
\usepackage[hyphens]{url} 
\usepackage{graphicx} 
\usepackage{natbib} 
\usepackage{caption} 
\usepackage{amsmath}
\usepackage{amssymb}
\usepackage{booktabs}
\usepackage{multirow}
\usepackage{xcolor}

\newcommand{\taum}{\boldsymbol{\tau}^{\mathrm{meas}}}

\title{FWBC-VLA: Force-Aware Whole-Body Compensation for Contact-Rich Loco-Manipulation}

\author{
    Yutian Zhang\textsuperscript{\rm 1}\textsuperscript{*},
    Siyuan Ma\textsuperscript{\rm 3}\textsuperscript{*},
    Liwen Yang\textsuperscript{\rm 1},
    Yang Li\textsuperscript{\rm 2},
    Ce Hao\textsuperscript{\rm 4},
    Haozhen Chi\textsuperscript{\rm 6},
    Dong Wei\textsuperscript{\rm 5}\textsuperscript{\textdagger},
    Qiaojun Yu\textsuperscript{\rm 2}\textsuperscript{\textdagger},
    Dibo Hou\textsuperscript{\rm 1}\textsuperscript{\textdagger}
}
\affiliations{
    \textsuperscript{\rm 1}Zhejiang University,
    \textsuperscript{\rm 2}Shanghai Artificial Intelligence Laboratory,
    \textsuperscript{\rm 3}Tsinghua University,
    \textsuperscript{\rm 4}Zhongguancun Academy\\
    \textsuperscript{\rm 5}Deep Robotics,
    \textsuperscript{\rm 6}Zhejiang University of Science and Technology\\
    \textsuperscript{*}Equal contribution. \textsuperscript{\textdagger}Corresponding authors.\\
    Project page: \url{https://ytydt-reuz.github.io/FWBC-VLA/}
}

\begin{document}

\maketitle

\begin{abstract}
  Contact-rich loco-manipulation requires a bridge between semantic action generation and physical interaction control.
  Existing Vision-language-action (VLA) models generate task-level actions from visual and linguistic observations, but cannot interpret the physical interactions induced by those actions.
  While the whole-body control (WBC) policy can stabilize the robot, it cannot distinguish task-relevant interaction forces from forces induced by external disturbances during manipulation.
  Although force/torque sensors provide direct measurements of physical interactions, retrofitting them entails additional hardware costs and substantial integration effort, particularly for platforms not designed with sensor integration in mind.
  To address this problem, we propose FWBC-VLA, a force-aware framework that bridges task-level VLA action generation and low-level whole-body compensation control for wheeled-legged robots.
  First, we introduce HSR-Force, a sensorless residual-torque estimator for inferring contact strength and its temporal variation.
  These contact estimates are then encoded as tokens and injected into the VLA action expert during action decoding, enabling the policy to perceive contact onset, sustained loading, and release.
  For loco-manipulation tasks, all parameters of the pretrained VLA backbone are fine-tuned on our WL\&Arm Dataset, which comprises more than 5,000 episodes.
  Moreover, the robot's proprioceptive state, the Jacobian-derived body-frame force estimate, and the estimated contact state are jointly fed into a compensation generator to produce corrective actions.
  The manipulation-centric actions are subsequently combined with the corrective actions and passed to the WBC policy for execution.
  Real-world experiments on whiteboard wiping and door opening with a door closer demonstrate the effectiveness of our FWBC-VLA in contact-rich loco-manipulation.
\end{abstract}

\section{Introduction}

\begin{figure}[!ht]
  \centering
  \includegraphics[width=\columnwidth]{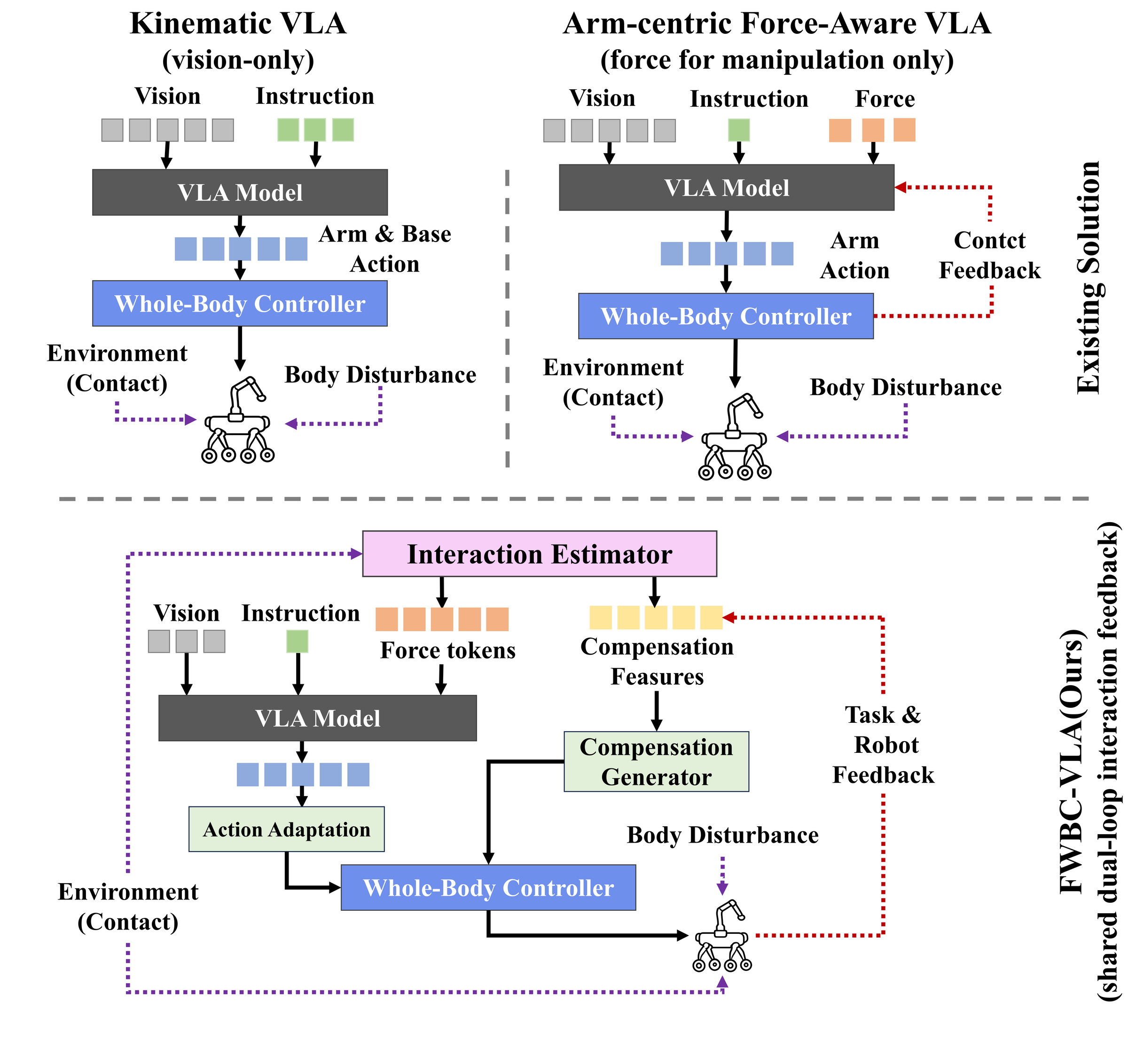}
  \caption{Comparison of VLA loco-manipulation paradigms. Kinematic whole-body VLA uses no explicit interaction feedback. Arm-centric force-aware VLA uses force for task adaptation but ignores body-level effects. We leverage a sensorless interaction estimator to construct dual feedback loops for both task actions and body compensation.}
  \label{fig:paradigms}
\end{figure}

Recent Vision-Language-Action (VLA) models have demonstrated significant potential to unify perception, language understanding, and action generation within an end-to-end policy \cite{brohan2023rt2,kim2024openvla}. 
As shown in Figure~\ref{fig:paradigms}, current VLA paradigms have been extended from fixed-base manipulation to whole-body loco-manipulation. Contact-rich loco-manipulation, however, requires explicit awareness of physical interactions\cite{grootn1,wholebodyvla2025,openhlm2026}. 
Contact-rich loco-manipulation, however, requires explicit awareness of physical interactions.
Arm-equipped wheeled-legged quadrupeds are especially appealing for this challenging task, offering greater stability than humanoids through their lower center of mass and four-legged support, while possessing enhanced terrain traversability relative to traditional wheeled robots.
Force applied on the end-effector (EE) can propagate through along the arm and perturb the body, Accordingly, task-oriented actions must be coordinated with whole-body stabilization.
This raises a central question: how can causal contact feedback be extracted and jointly utilized by the VLA policy and whole-body control (WBC)?

A common solution is to maintain a modular separation between task-level action generation and body stabilization.
While this design offers practical advantages, the manipulation policy and controller merely exchange kinematic commands, resulting in insufficient closed-loop feedback about contact onset, sustained loading, and release.
End-to-end whole-body VLAs avoid such explicit handoffs, but they require substantial whole-body demonstration data and typically capture physical interactions only implicitly through vision and proprioception~\cite{wholebodyvla2025}.
In both cases, the robot may execute the intended motion while remaining unaware of how task related contact forces evolve and perturb the mobile base.
Contact-aware VLA methods~\cite{Hou2024AdaptiveCP,forcevla2025,forcevla2} demonstrate that explicit force/torque (F/T) feedback improves performance in contact-rich manipulation.
Nevertheless, most of these approaches are arm-centric and rely on dedicated F/T sensors.
Unfortunately, many existing robotic platforms lack integrated F/T sensors, and retrofitting introduces mechanical, electrical, and calibration challenges. 
This call for a sensorless contact-feedback interface to estimate physical interactions from the robot’s proprioceptive signals, provides contact information to the VLA policy, and converts estimated arm loads and body posture deviations into corrective commands for the WBC.

To realize this interface, we propose FWBC-VLA, a sensorless, force-aware VLA framework for arm-equipped wheeled-legged quadrupeds.
Our History-State Residual Force Estimator (HSR-Force) employs dual LSTM networks to improve force estimation accuracy under both dynamic and static conditions by synchronizing arm, leg, and base proprioception.
It preserves directional six-dimensional residual joint torques and summarizes its magnitudes and temporal trend into contact descriptor. 
Short interaction histories are further encoded into force tokens and late-fused into the action expert, enabling the VLA policy to perceive force evolution and modulate loco-manipulation action generation according to varying physical contacts.
Furthermore, estimated external joint torques are transformed through respective Jacobians into Cartesian wrench estimates at the EE and base. 
Combined with base proprioception, these wrenches are adopted to assess base deviations and contact evidence, triggering a compensation action generator to keep the robot stable during the force-interaction task.
Our contributions are summarized as follows:
\begin{itemize}
  \item We propose FWBC-VLA, a framework with force-aware interface that converts proprioceptive signals into shared interaction for VLA action and whole-body compensation.
  \item We present HSR-Force for contact detection and force estimation, establishing a causal physical feedback channel without requiring dedicated physical sensors.
  \item We develop a hierarchical execution strategy, in which interaction representations simultaneously adapt manipulation actions and mitigate body-level disturbances.
\end{itemize}

\section{Related Work}

\paragraph{Contact-aware Vision-language-action Models.}
Recently, VLA systems~\cite{brohan2023rt2,kim2024openvla,openpi0,openpi0.5} have attracted significant attention because of their strong generalization and manipulation capabilities.
Representative works that incorporate force sensing~\cite{tavla2025,forcevla2,fdvla2026} enhance manipulation stability and precision, yet they typically treat force as an additional feedback signal for arm control and overlook its effects on the robot body.
Tactile-based methods~\cite{torlvla2026,tla,tac-man} and modality-fusion approaches~\cite{Huang2025TactileVLA,impact,should} improve robustness under occlusion, but they cannot fully replace direct force sensing because force estimates derived from tactile signals remain indirect and susceptible to noise.
Nevertheless, existing approaches remain limited in their ability to incorporate force feedback into WBC.

\paragraph{Sensorless force and contact estimation.}
External forces are typically estimated by modeling robot dynamics, with discrepancies between predicted and measured signals serving as indicators of external interactions. Traditional methods \cite{biact,yamane2026,shi2026minimalistcompliancecontrol} require extensive system modeling and engineering effort but often struggle with actuator imperfections such as dead zones \cite{inami2024lossfunctionconsideringdead} and torque ripple \cite{zhu2025cycloidal}. Learning-based approaches reduce the dependence on explicit dynamics modeling but introduce new limitations. Supervised methods require paired real-world or simulated contact data \cite{supervised_anomalydetection,liang2021contact}, while unsupervised methods such as autoencoder-based anomaly detection are mainly designed for contact detection rather than continuous force estimation \cite{unsupervised_anomalydetection}. 
Meanwhile, inverse-dynamics learning remains dependent on robot-specific models or high-end robotic platforms \cite{zhu2025cycloidal,yilmaz2020neural}. 
Despite these advances, accurate force estimation remains challenging during dynamic whole-body motions with complex interactions.

\paragraph{Whole-body Loco-manipulation Controllers.}
To move beyond isolated manipulation or base motion, VLA-based whole-body policies have been proposed \cite{momanipvla2025,sgvla2026,wholebodyvla2025}.
To merge perception, planning, and decision-making, QUAR-VLA \cite{quarvla2023} integrates visual information and instructions to generate actions.
ODYSSEY \cite{odyssey2025} combines VLM for task planning with WBC to control quadruped robots equipped with manipulators.
Prior work \cite{leggedforce2024} has proposed a learned WBC policy that commands the manipulator, while automatically adjusting the robot's body to achieve desired position and force targets.
\cite{unifiedlegged2025} incorporates an explicit manipulator kinematic model into the RL framework, providing feedback on how body postures map to the manipulator's workspace to guide exploration and mitigate the local optimum problem.
Although these methods improve WBC through perception, planning, kinematic modeling, or learned control, they lack an explicit mechanism to incorporate force feedback for contact-rich loco-manipulation tasks.

\section{WL\&Arm Dataset}
\subsection{Force-Intent WBC Teleoperation}
Although VR teleoperation has become a mainstream approach for data collection, 
contact-aware datasets for wheeled-legged quadruped loco-manipulation remain absent. 
We aim to address this limitation with a feasible solution.

We identify two core challenges. 
First, conventional teleoperation records only motion without capturing the operator's force intent, which complicates force-aware supervision.
Second, wheeled-legged robots differ from embodiments adopted to pretrain most VLA models, creating mismatches in both the base and manipulator action spaces.

To address these challenges, we develop a Pico-based hybrid position-force teleoperation system.
As shown in Figure~\ref{fig:wlarm-dataset}, the controller specifies tangential and normal EE force commands, which are applied as contact feedforward terms and logged as force-intent annotations.
To avoid command-intent leakage, these annotations are used only as dataset metadata and inspection references, never as inputs to the estimator, VLA policy, or compensation module.
This improves the quality of contact-rich data collected from tasks including whiteboard wiping and door opening with a door closer.
To reduce embodiment mismatch, we adopt base velocity and steering angle as action outputs rather than joint-specific commands, which decouples VLA action prediction from the wheel-leg controller while preserving execution by the joint-level controller.

\subsection{Statistics of the WL\&Arm Dataset}
\begin{figure}[htb]
  \centering
  \includegraphics[width=\columnwidth]{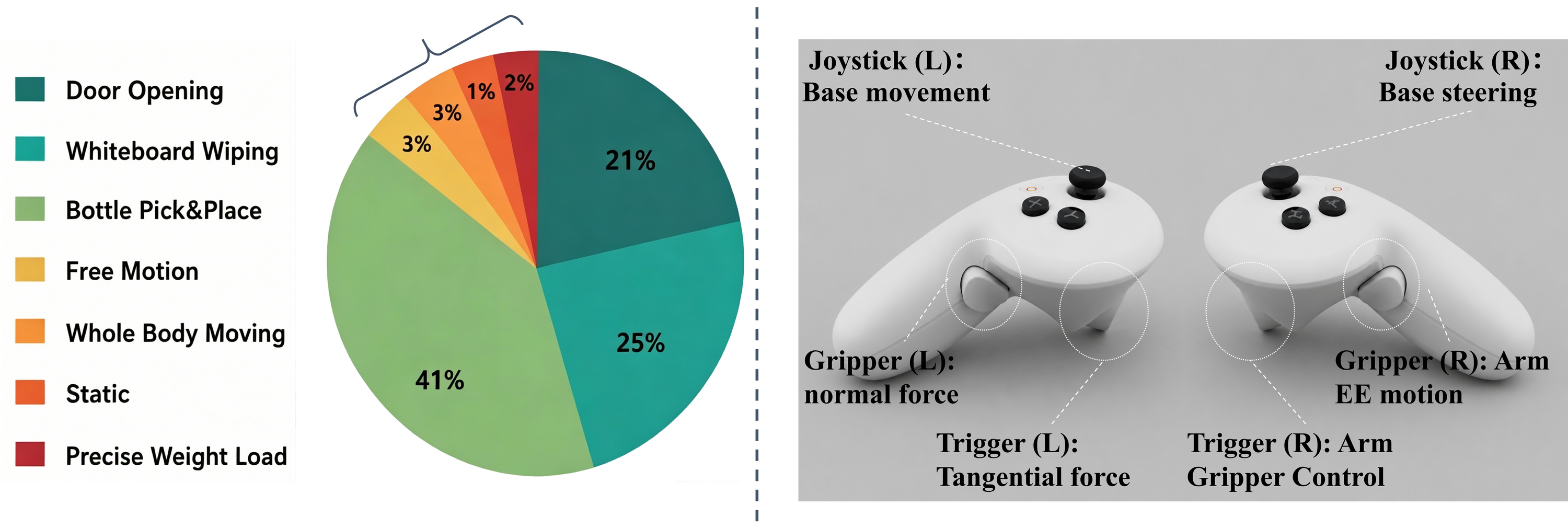}
  \caption{Illustrations of dataset and data collecting. Left: Composition of the WL\&Arm dataset. Right: Pico-based teleoperation method with force commands.}
  \label{fig:wlarm-dataset}
\end{figure}

Our Wheel-Legged (WL) \&Arm Dataset, consisting of more than 5,000 teleoperation episodes, will be released publicly.
Figure~\ref{fig:wlarm-dataset} summarizes the composition of WL\&Arm. 
Our dataset provides 200-Hz joint data supporting for contact estimation, alongside 15-Hz loco-manipulation demonstrations.
The three principal task subsets are bottle pick-and-place (41\%), whiteboard wiping (25\%), 
and door opening (21\%).  
Force calibration data includes applied loads of 0.36 kilograms and 0.72 kilograms, prodeced by a force gauge, These measurements are adopted to train and evaluate the performance under standard conditions.

\section{Method}

\begin{figure*}[t]
  \centering
  \includegraphics[width=\textwidth]{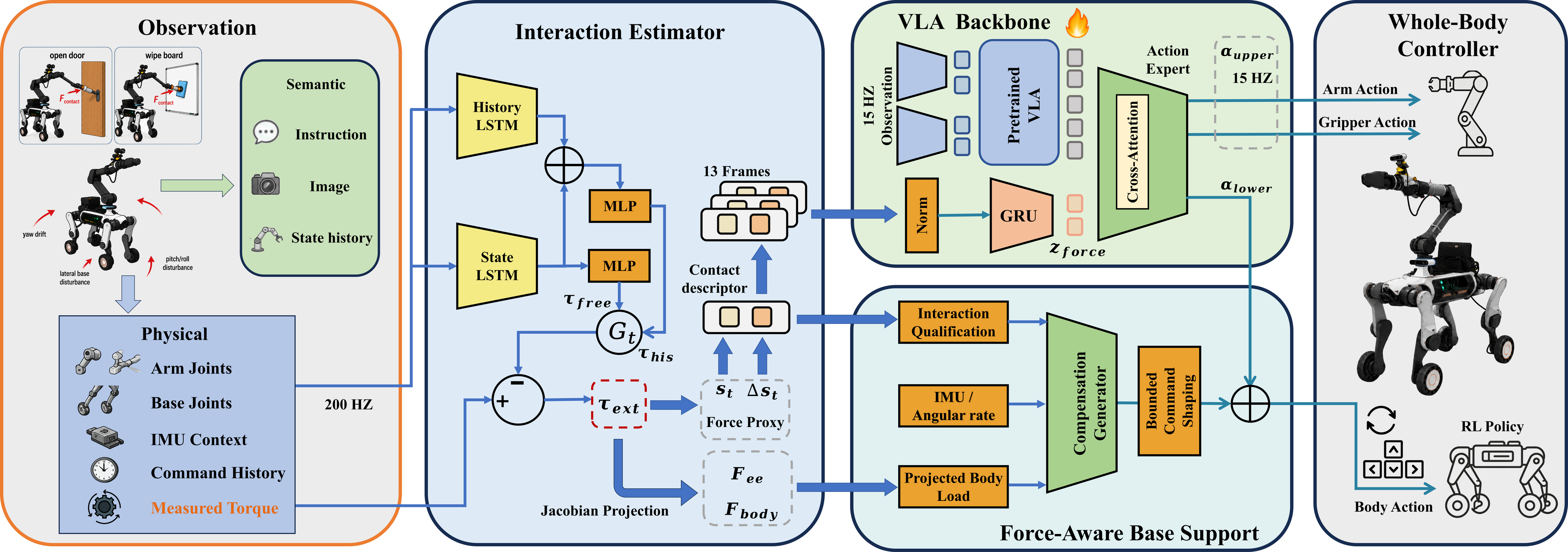}
  \caption{FWBC-VLA framework introduces a sensorless, force-aware interface between the VLA and WBC policies. The estimator transforms proprioceptive states into causal interaction representations, which are shared across two branches: interaction-conditioned action generation for VLA, and task-relevant body compensation for WBC.}
  \label{fig:framework}
\end{figure*}

Figure~\ref{fig:framework}  summarizes FWBC-VLA with three components:
sensorless interaction estimation, interaction-conditioned VLA action
generation, and whole-body compensation.
The estimated interaction representation is used by two complementary
pathways: it conditions task-level action generation in the VLA and
provides physical feedback for base correction.
The downstream WBC remains responsible for balance regulation and
low-level command tracking.

\subsection{Sensorless Interaction Estimation}

Let $\taum_t \in \mathbb{R}^{n_a}$ denote the measured
arm-joint torque. Rather than directly estimating a calibrated external wrench, we
estimate a joint-space interaction residual by decomposing the measured
torque as:
\begin{equation}
  \taum_t = \boldsymbol{\tau}_{\mathrm{free},t}
  + \boldsymbol{\tau}_{\mathrm{ext},t}
  + \boldsymbol{\epsilon}_t,
\end{equation}
where $\boldsymbol{\tau}_{\mathrm{free},t}$ represents torque caused by robot motion and internal dynamics, $\boldsymbol{\tau}_{\mathrm{ext},t}$ captures environment-induced interaction residuals, and $\boldsymbol{\epsilon}_t$ accounts for unmodeled effects. Note that $\boldsymbol{\tau}_{\mathrm{ext},t}$ is treated as a latent interaction variable rather than a directly calibrated force measurement.

The estimator contains two compact experts that operate synchronously at 200 Hz and predict the free-motion torque. The history expert uses a causal torque-and-motion history to suppress free-motion prediction noise. The state expert instead uses the current arm, base, leg, and IMU state but does not read torque history, reducing the risk that sustained contact is absorbed into the free-motion prediction. We denote their outputs by $\hat{\boldsymbol{\tau}}_{\mathrm{free,hist},t}$ and $\hat{\boldsymbol{\tau}}_{\mathrm{free,state},t}$. The state residual and fixed gate are
\begin{align}
  \mathbf{r}_{\mathrm{state},t}
  &= \taum_t-\hat{\boldsymbol{\tau}}_{\mathrm{free,state},t},\\
  \alpha_t
  &=G_{\mathrm{fixed}}\!\left(\|\mathbf{r}_{\mathrm{state},t}\|_2\right),
  \qquad \alpha_t\in[0,1].
\end{align}
The fixed gate combines the two free-motion estimates:
\begin{align}
  \hat{\boldsymbol{\tau}}_{\mathrm{free},t}
  &=(1-\alpha_t)\hat{\boldsymbol{\tau}}_{\mathrm{free,hist},t}
    +\alpha_t\hat{\boldsymbol{\tau}}_{\mathrm{free,state},t},\\
  \hat{\boldsymbol{\tau}}_{\mathrm{ext,raw},t}
  &=\taum_t-\hat{\boldsymbol{\tau}}_{\mathrm{free},t}.
\end{align}
Thus, $\alpha_t\approx0$ favors the low-noise history expert during free motion, whereas $\alpha_t\approx1$ favors the state expert during interaction. Both experts and the gate belong to one 200-Hz estimation chain; no slower estimator or outer mixing gate is used.

We fit the free-torque experts on free-motion windows, for which $\boldsymbol{\tau}_{\mathrm{ext},t}\approx\mathbf{0}$. Reviewed contact-hold and transition frames are masked whenever phase annotations are available so that sustained interaction is not learned as ordinary dynamics. All six residual channels are retained to preserve directional information for downstream Jacobian projection.

\subsection{Causal Physical Interaction Representation}
We summarize the raw six-joint residual using a compact two-channel
contact descriptor:
\begin{align}
s_t &=
\left\|
\hat{\boldsymbol{\tau}}_{\mathrm{ext,raw},t}
\right\|_2,
\\
\Delta s_t &= s_t - s_{t-1},
\\
\mathbf{d}^{\mathrm{int}}_t
&=
\begin{bmatrix}
s_t & \Delta s_t
\end{bmatrix}^{\top}
\in \mathbb{R}^{2}.
\end{align}
Here, $s_t$ represents the estimated interaction strength proxy, whereas $\Delta s_t$ captures its local temporal trend, distinguishing increasing loading from unloading or release. We interpret
$\mathbf{d}^{\mathrm{int}}_t$ as a proxy-valued physical interaction state rather than a calibrated wrench. Its scale therefore remains dependent on the residual estimator and the sampling configuration.

For directional compensation, damped least-squares Jacobian inversion maps
the joint-torque residual to an equivalent six-dimensional interaction
wrench ${}^{\mathcal E}\hat{\mathbf w}_{\mathcal E,t}$
\cite{haddadin2017robot}.
Here, $\mathcal E$ is the EE control frame used for Jacobian
computation, and $\mathcal B$ is the robot-body frame at the body origin.
The same physical interaction is re-expressed about the body origin as
${}^{\mathcal B}\hat{\mathbf w}_{\mathcal B,t}$, including the lever-arm
moment. Both representations serve
as directional and relative-load features.
Details are provided in Appendix.


\subsection{Interaction-Conditioned Action Generation}

To provide the VLA with causal physical feedback, we encode a short history of the interaction representation:
\begin{equation}
  \mathbf{D}^{\mathrm{int}}_t=\left[\mathbf{d}^{\mathrm{int}}_{t-K+1},\ldots,\mathbf{d}^{\mathrm{int}}_t\right]\in \mathbb{R}^{K \times 2},
\end{equation}
where$\mathbf{d}^{\mathrm{int}}_i=[s_i,\Delta s_i]^\top$.
The implemented configuration uses $K=13$ at 200 Hz. The 13 samples span 60 ms between the first and last timestamp and capture the most recent interaction trend without using future measurements. 
At the action-expert width, a normalization layer followed by a GRU temporal encoder produces the force tokens as:
\begin{equation}
  \mathbf{Z}^{\mathrm{int}}_t=E_{\mathrm{int}}\left(\mathbf{D}^{\mathrm{int}}_t\right)
\in \mathbb{R}^{N_{\mathrm{int}} \times 1024}.
\end{equation}
The GRU only encodes interaction history and does not serve as an additional action decoder.

We instantiate the VLA policy using the pretrained $\pi_{0.5}$
backbone \cite{openpi0.5}. 
The policy operates at 15 Hz, whereas HSR-Force produces residual-torque estimates at 200 Hz.
For each VLA query timestamp $t$, we retrieve the $K=13$ most
recent contact descriptors whose timestamps are not later
than $t$.
At 200 Hz, the 13 samples cover 60 ms between the earliest and
latest timestamps. We adopt a latest-before-query synchronization
rule and never use interaction measurements later than the current
policy timestamp, thereby preserving causal deployment.
The pretrained VLA backbone is fully fine-tuned on the target embodiment data. Within its Gemma action expert, visual-language prefix tokens $T_{prefix}$ and noisy action tokens $T_{action}$ are processed jointly. Force is fused into the action hidden states rather than propagated through the visual backbone. Specifically, action hidden states provide the cross-attention queries and force tokens provide keys and values:
\begin{equation}
  \widetilde{\mathbf{Z}}^{action}_t = \mathbf{Z}^{action}_t
  + \alpha_f g_t \operatorname{CA}(\mathbf{Z}^{action}_t,\mathbf{Z}^{force}_t),
\end{equation}
where $\operatorname{CA}$ is cross-attention, $g_t$ is a learned force-presence gate, and $\alpha_f$ scales the force residual. $Z^{\mathrm{force}}_t$ in Equation~(11) refers to the force tokens defined in Equation~(10).
This fusion is applied to the action expert's final hidden states before the single output projection, which predicts the flow velocity:
\begin{align}
  (\mathbf{Z}^{prefix},\mathbf{Z}^{action})
  &=\operatorname{VLA}_{\theta}(T_{prefix},T_{action}),\\
  \mathbf{v}_t&=\operatorname{action\_out\_proj}
  (\widetilde{\mathbf{Z}}^{action}_t).
\end{align}
There is therefore one Action Expert (Action Decoder), not a serial Action Expert--Action Decoder pair. The conditioned policy predicts an action chunk
\begin{equation}
  \pi_\theta(\mathbf{I}_t,\ell,\mathbf{x}_t,\mathbf{F}_t)
  \rightarrow
  \mathbf{A}_{t:t+H-1}.
\end{equation}
Here $\mathbf{A}_{t:t+H-1}$ contains arm, gripper, and nominal base actions. 
The image-language context remains unchanged. The interaction interface is introduced only as a physical feedback pathway and does not require auxiliary contact classification or phase prediction objectives.

\subsection{Whole-Body Compensation Generation}
The second output pathway of the physical interaction interface provides whole-body compensation outside the VLA action expert. 
This branch uses the gripper state and interaction descriptor as gating cues to qualify whether a disturbance is task-relevant, and maps the current load estimate and posture deviation to a bounded base-velocity residual.
The downstream WBC remains responsible for balance and low-level tracking.

The compensation generator receives the projected body-frame load
proxy $\hat{F}_{\mathrm{body},t}$, the IMU orientation and angular rate,
and the nominal base action $u^{\mathrm{base,nom}}_t$.
These inputs respectively describe the transmitted interaction load,
the resulting body deviation, and the original task-level motion
intent. $\hat{F}_{\mathrm{body},t}$ is derived from the manipulator residual and
does not represent an independent force measurement on the chassis. The posture deviation is defined as follows:
\begin{equation}
    \Delta\boldsymbol{\eta}^{imu}_t
    =
    \left[
    \boldsymbol{\eta}^{imu}_t-\boldsymbol{\eta}^{imu,ref}_t,\,
    \boldsymbol{\omega}^{imu}_t
    \right],
\end{equation}
where $\boldsymbol{\eta}^{imu,ref}_t$ is the recent free-motion body orientation reference. A compact multilayer perceptron predicts an
unconstrained residual proposal:
\begin{equation}
    \bar{\mathbf{r}}^{base}_t=
    C_\psi\!\left(
    \hat{\mathbf{F}}_{body,t},
    \Delta\boldsymbol{\eta}^{imu}_t,
    \mathbf{u}^{base,nom}_t
    \right),
\end{equation}
where $\bar{\mathbf{r}}^{base}_t\in\mathbb{R}^{3}$ corresponds to residual planar velocity and yaw rate $(\Delta v_x,\Delta v_y,
\Delta\omega_z)$.

During training, the compensation target is defined as the difference between the task-level nominal base command and the executed force-supported base command:
\begin{equation}
    \mathbf{r}^{base,*}_t
    =
    \mathbf{u}^{base,exec}_t
    -
    \mathbf{u}^{base,nom}_t .
\end{equation}
The compensation generator is trained with an $\ell_1$ regression objective on this residual target. The final residual is bounded
before execution:
\begin{equation}
    \mathbf{r}^{base}_t
    =
    \mathbf{r}_{\max}\odot
    \tanh\!\left(\bar{\mathbf{r}}^{base}_t\right).
\end{equation}
The executed base command is
\begin{equation}
    \mathbf{u}^{base}_t =
    \mathbf{u}^{base,nom}_t
    + \mathbf{r}^{base}_t .
\end{equation}
Bounded command shaping applies a deadband, low-pass filtering, slew-rate limits, and final command-headroom clipping before the residual is sent to the WBC. Stale force estimates or invalid IMU/load signals set the residual to zero, preserving a bounded interface to the robot controller.

\section{Experiments}
In this section, We evaluate the estimation, compensation, and loco-manipulation performance of FWBC-VLA on diverse contact-rich tasks in the real world.
The experiments address the following research questions:
\begin{itemize}
\item \textbf{Q1:} How accurately can HSR-Force estimate external interaction without an F/T sensor?
\item \textbf{Q2:} How do the state and history heads and their fixed-gate fusion affect estimation performance?
\item \textbf{Q3:} How does FWBC-VLA perform in real-world contact-rich loco-manipulation tasks, and what advantages does force-aware control offer?
\item \textbf{Q4:} Which components contribute most significantly to the performance improvements of FWBC-VLA?
\end{itemize}

\subsection{Force-Estimator Evaluation}
\label{sec:force-estimator}
\subsubsection{Sensorless Estimation Baselines.}
We compare HSR-Force with NEXT, a learned external-force estimator \cite{Oh2026FACTR2L}; GMO-SI, an identified generalized momentum observer \cite{haddadin2017robot}; and DF-MLP, a direct proprioception-to-force multilayer perceptron \cite{shan2023sensorless}.
NEXT, GMO-SI, and HSR-Force use the same residual-to-force adapter, whereas DF-MLP uses its direct scalar-force output.
DF-MLP does not output external joint torque and is therefore omitted from the torque-residual metrics.
All trainable estimators are trained on our 200-Hz data using five different seeds.
We additionally evaluate scalar-force accuracy on held-out static and dynamic recordings with known payloads of 0.36 and 0.72~kg. Their ground-truth loads are computed as $F^{\mathrm{GT}}=mg$, corresponding to 3.53 and 7.06~N, respectively.

\subsubsection{Estimation Metrics.}
We evaluate force estimation using N-MAE, torque residual errors, and the AUC for contact detection. 
N-MAE reports the scalar force MAE in newtons. S/M denotes static/moving. 
Free and Dynamic $\tau$ MAE average $|\hat{\tau}^{\mathrm{ext}}_{t,j}|$ over all free-motion frames or only dynamic free-motion frames, respectively. $\tau$ R90 is the median episode-wise 90th percentile of $\|\hat{\boldsymbol{\tau}}^{\mathrm{ext}}_t\|_2$ for static/dynamic (S/D) motion. AUC denotes the area under the receiver operating characteristic curve, computed by sweeping the contact-score threshold.
Touch AUC measures contact-versus-free discrimination around touch events, whereas Door-phase AUC measures the same discrimination across door-manipulation phases.
Each metric is evaluated on 25 manually reviewed episodes.

\subsubsection{Overall performance on estimation (Q1).}
As summarized in Table~\ref{tab:force-evaluation}, HSR-Force achieves the lowest zero-load force error and the highest Touch AUC, while also reducing the static and dynamic tail residuals. 

\begin{table*}[t]
\centering
\small
\renewcommand{\arraystretch}{0.92}
\setlength{\tabcolsep}{2.5pt}
\begin{tabular}{@{}lccccccc@{}}
\toprule
& \multicolumn{3}{c}{Zero-load evaluation}
& \multicolumn{2}{c}{Contact discrimination}
& \multicolumn{2}{c}{Known-load force evaluation} \\
\cmidrule(lr){2-4}\cmidrule(lr){5-6}\cmidrule(lr){7-8}
Method &
\shortstack{Force MAE\\(N) $\downarrow$} &
\shortstack{All-free $\tau$ MAE\\(N\,m) $\downarrow$} &
\shortstack{$\tau$ R90 S/D\\(N\,m) $\downarrow$} &
\shortstack{Touch\\AUC $\uparrow$} &
\shortstack{Door-phase\\AUC $\uparrow$} &
\shortstack{Frame MAE S/D\\(N) $\downarrow$} &
\shortstack{Episode MAE S/D\\(N) $\downarrow$} \\
\midrule
\multicolumn{8}{@{}l}{\emph{Related-method baselines}} \\
NEXT & 0.28 & 0.66 & 1.26/3.37 & 0.95 & 0.82 & 2.52/1.43 & 2.57/1.53 \\
GMO-SI & 0.43 & 0.56 & 1.14/2.97 & 0.94 & 0.83 & 1.72/1.59 & 0.78/1.18 \\
DF-MLP & 0.21 & -- & -- & 0.47 & 0.50 & 2.01/1.87 & 1.98/1.26 \\
\textbf{HSR-Force} & \textbf{0.15} & \textbf{0.37} & \textbf{0.64}/\textbf{2.79} & \textbf{0.97} & \textbf{0.85} & \textbf{1.09}/\textbf{1.21} & \textbf{0.28}/\textbf{1.06} \\
\midrule
\multicolumn{8}{@{}l}{\emph{Shared dual-head ablation (mean with s.d. in parentheses; three seeds)}} \\
State only
& $1.13(.35)$ & $0.82(.04)$ & $1.47(.16)$/$3.46(.32)$
& $0.92(.02)$ & $0.80(.01)$
& $1.99(.67)$/$2.25(.80)$ & $1.08(.45)$/\textbf{$1.70(.45)$} \\
History only
& $0.93(.24)$ & \textbf{$0.60(.03)$} & \textbf{$1.23(.59)$}/\textbf{$2.94(.15)$}
& $0.87(.03)$ & $0.78(.01)$
& $2.33(1.03)$/$2.09(.62)$ & $1.93(.99)$/$1.98(.66)$ \\
Ours (Dual\&gate)
& \textbf{$0.25(.13)$} & $0.52(.24)$ & \textbf{$1.23(.59)$}/$2.96(.17)$
& \textbf{$0.94(.03)$} & \textbf{$0.83(.02)$}
& \textbf{$1.51(.42)$}/\textbf{$2.02(.81)$}
& \textbf{$0.81(.53)$}/$1.84(.78)$ \\
\bottomrule
\end{tabular}
\caption{Unified force-estimator results. S/D denotes static/dynamic; ablation entries are mean (s.d.) over three seeds. Boldface marks the best result within each method group; dashes denote unavailable metrics.}
\label{tab:force-evaluation}
\end{table*}

\begin{figure}[htb]
  \centering
  \includegraphics[width=\columnwidth]{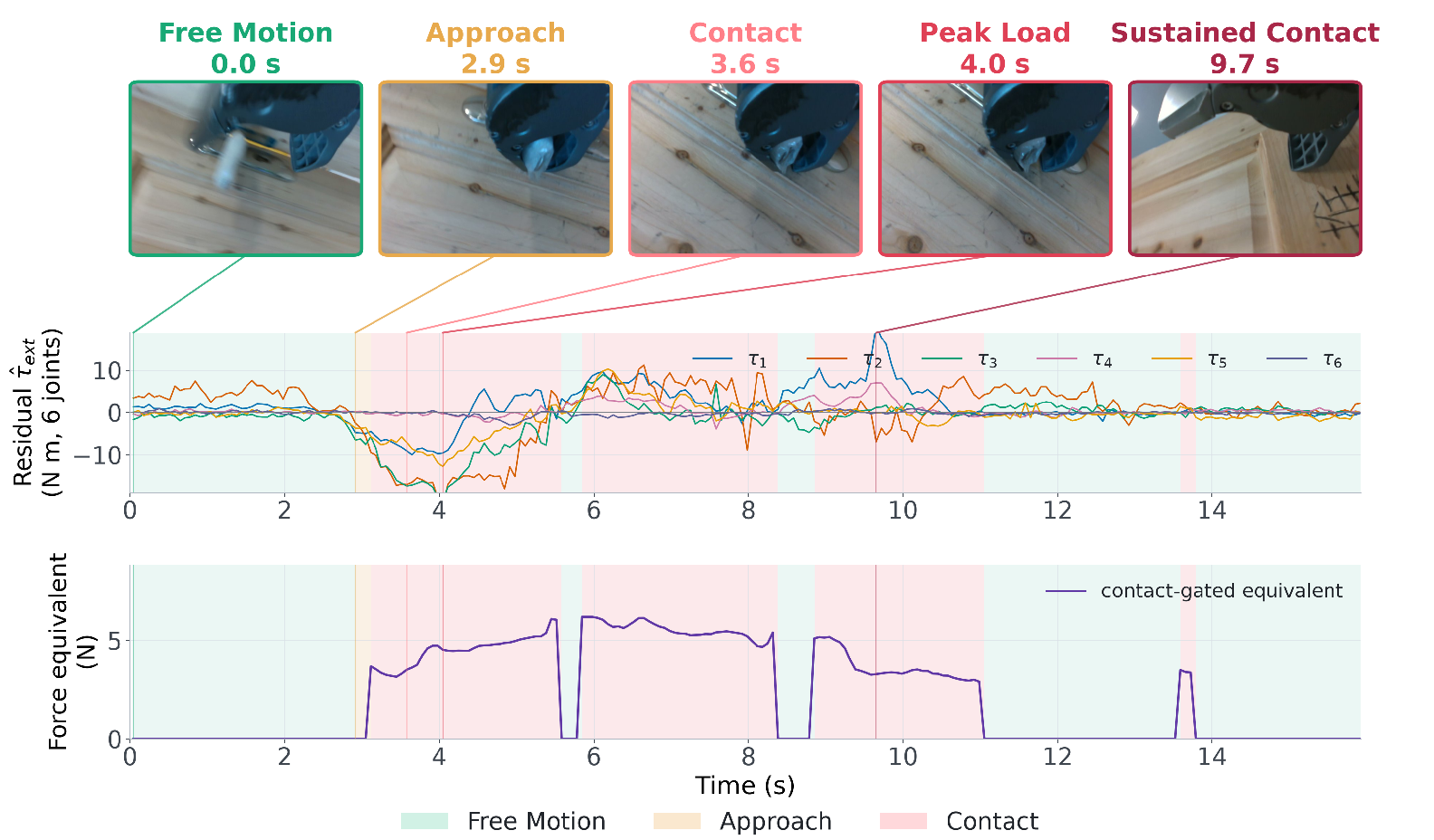}
  \caption{Video-aligned contact estimation of door opening.}
  \label{fig:door-force-estimation}
\end{figure}

The video-aligned result further shows that HSR-Force responds near contact onset and remains active during sustained door loading (Fig.~\ref{fig:door-force-estimation}).

\subsubsection{Shared Dual-Head Ablation (Q2).}
We ablate the state head, history head, and fixed-gate dual-head fusion using the same causal context.
Torque history reduces free-motion residual error, particularly during motion, whereas the history-free state head is more sensitive to contact onset. Their fixed-gate fusion yields the best Touch and Door-phase AUCs (Table~\ref{tab:force-evaluation}). 

On the non-zero known-load experiments, HSR-Force achieves the lowest baseline frame Force MAE under both static and dynamic recordings and the lowest dynamic error, whereas GMO-SI gives the lowest static error. 
Within the three-seed ablation, fixed-gate fusion provides the best known-load errors overall while retaining competitive zero-load residual errors and the best contact-discrimination AUCs.
Together, the zero-load and known-load experiments provide complementary physical references.

\begin{table}[t]
\centering
\small
\renewcommand{\arraystretch}{0.92}
\setlength{\tabcolsep}{1.2pt}
\begin{minipage}[t]{0.49\columnwidth}
\centering
\begin{tabular}{@{}lccccc@{}}
\toprule
\multicolumn{6}{c}{\textbf{Whiteboard Wiping}} \\
Method & S1 & S2 & S3 & S4 & S5 \\
\midrule
\multicolumn{6}{@{}l}{\emph{Without force}} \\
OpenVLA     & 60 & 48 & 4  & 0  & 0  \\
StarVLA     & 64 & 56 & 12 & 0  & 0  \\
OpenPI 0.5      & 80 & 28 & 20 & 12 & 12 \\
GR00T N1.6  & 76 & 36 & 8  & 0  & 0  \\
\midrule
\multicolumn{6}{@{}l}{\emph{With force}} \\
ACP         & 72 & 64 & 44 & 20 & 20 \\
ForceVLA    & 84 & 68 & 36 & 24 & 24 \\
FWBC-GT    & 84 & 84 & 76 & 60 & 60 \\
\textbf{FWBC-VLA} & \textbf{88} & \textbf{84} & \textbf{76} & \textbf{64} & \textbf{64} \\
\bottomrule
\end{tabular}
\end{minipage}\hfill
\begin{minipage}[t]{0.49\columnwidth}
\centering
\begin{tabular}{@{}lccccc@{}}
\toprule
\multicolumn{6}{c}{\textbf{Door Opening}} \\
Method & S1 & S2 & S3 & S4 & S5 \\
\midrule
\multicolumn{6}{@{}l}{\emph{Without force}} \\
OpenVLA     & 16 & 8  & 0  & 0  & 0  \\
StarVLA     & 20 & 12 & 0  & 0  & 0  \\
OpenPI 0.5  & 80 & 40 & 12 & 0  & 0  \\
GR00T N1.6  & 76 & 32 & 8  & 0  & 0  \\
\midrule
\multicolumn{6}{@{}l}{\emph{With force}} \\
ACP         & 72 & 44 & 12 & 4  & 4  \\
ForceVLA    & 76 & 56 & 24 & 12 & 12 \\
FWBC-GT    & 88 & 84 & 64 & 48 & 48 \\
\textbf{FWBC-VLA} & \textbf{92} & \textbf{84} & \textbf{64}& \textbf{52} & \textbf{52} \\
\bottomrule
\end{tabular}
\end{minipage}
\caption{Stage success rates (\%; 25 trials/task). S1--S5 follow setup in Sec.~\ref{sec:real-world}; rows are grouped by force input.}
\label{tab:real-world-comparison}
\end{table}

\begin{table}[t]
    \centering
    \small
    \renewcommand{\arraystretch}{0.92}
    \setlength{\tabcolsep}{1.2pt}
        \begin{tabular}{@{}lccrrrrr@{}}
            \toprule
            Method
            & FI
            & BC
            & \shortstack{Handle\\Press}
            & \shortstack{Push\\w/o closer}
            & \shortstack{Push\\w closer}
            & \shortstack{Board\\Cleaned}
            & Avg. \\
            \midrule
            w/o Force
            & --
            & --
            & 28
            & 12
            & 0
            & 8
            & 12.0 \\

            FI only
            & $\checkmark$
            & --
            & \textbf{52}
            & 56
            & 0
            & 32
            & 35.0 \\

            FWBC-VLA
            & $\checkmark$
            & $\checkmark$
            & \textbf{60}
            & \textbf{72}
            & \textbf{52}
            & \textbf{76}
            & \textbf{65} \\
            \bottomrule
        \end{tabular}
    \caption{Component ablation. Entries are contact-stage success rates (\%); Avg.\ is the unweighted mean. }
    \label{tab:comp_ablation}
\end{table}

\begin{figure}[!b]
  \centering
  \includegraphics[width=\columnwidth]{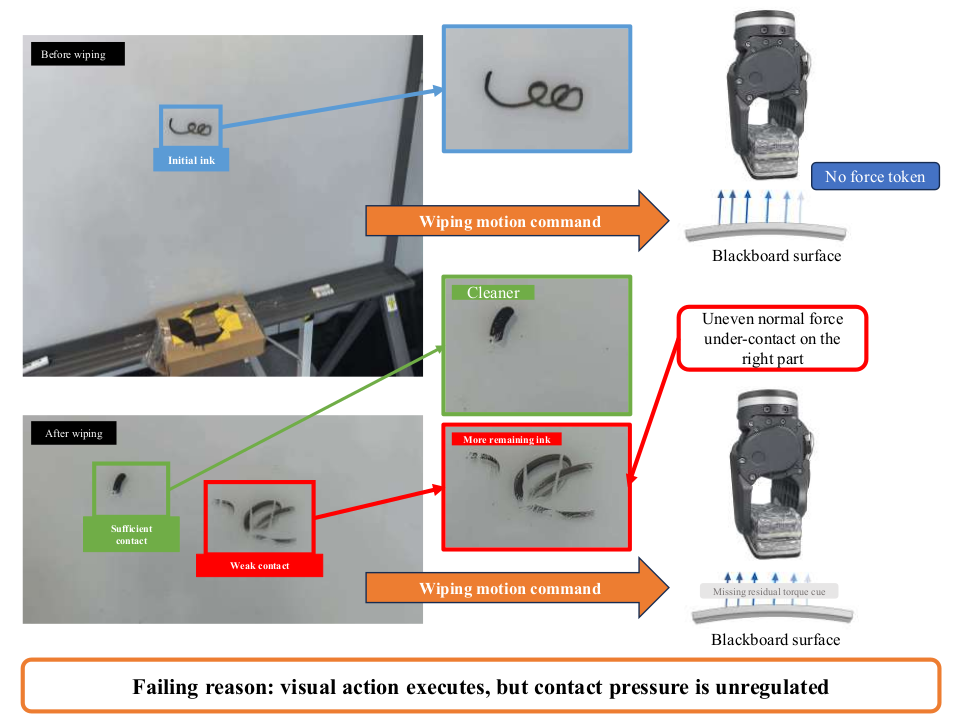}
  \caption{Effect of removing force conditioning during whiteboard wiping.}
  \label{fig:force-effect}
\end{figure}

\begin{figure*}[t]
  \centering
  \includegraphics[width=\textwidth]{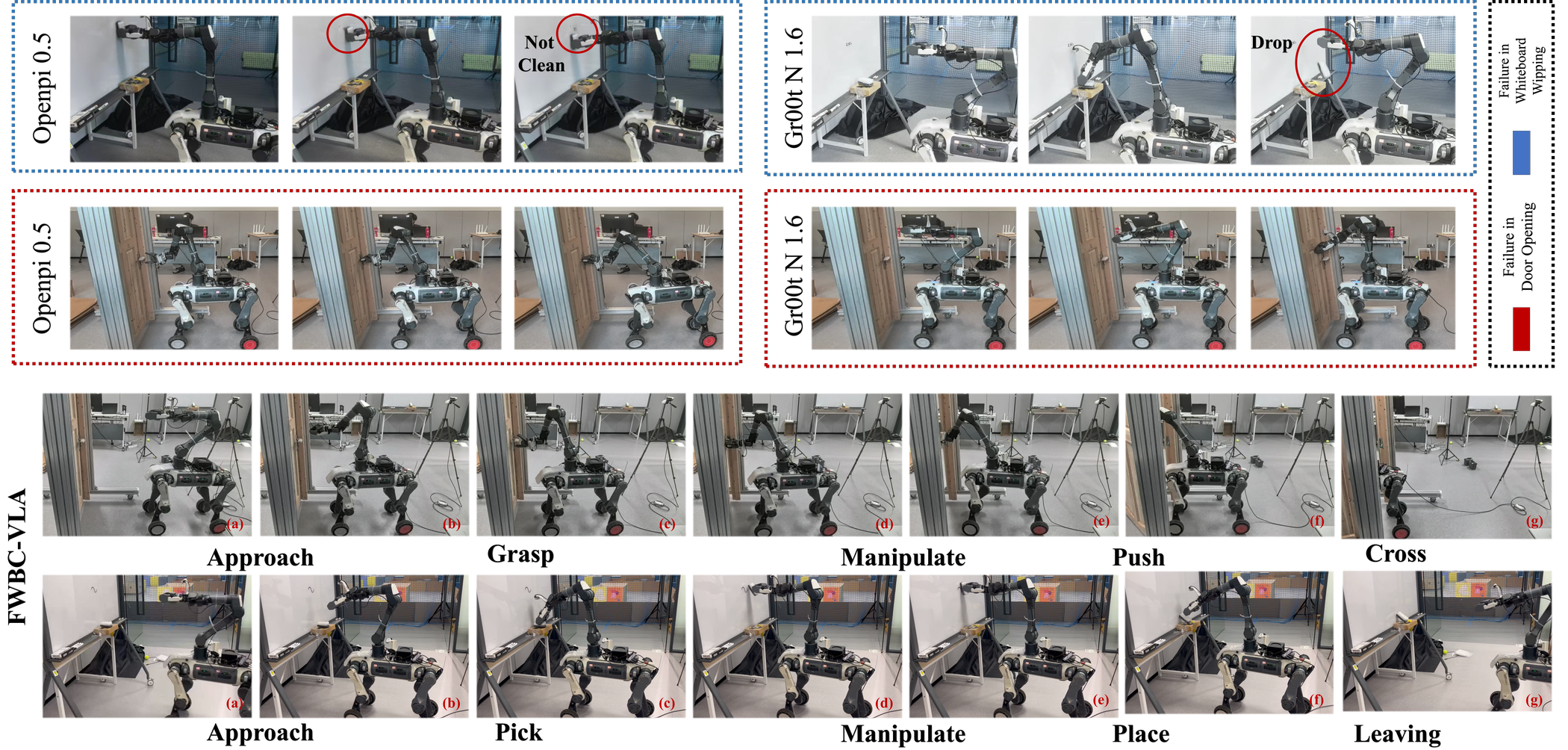}
  \caption{Representative real-world rollouts. Baseline failures on both tasks (top). FWBC-VLA succeeds in whiteboard wiping and door opening (bottom).}
  \label{fig:real-world-rollouts}
\end{figure*}

\subsection{Experiments in the Real World}
\label{sec:real-world}

\subsubsection{Setup and evaluation metrics.}
Experiments are conducted on a DeepRobotics M20S wheeled-legged quadruped equipped with a CM1 6-DoF robotic arm and a 1-DoF gripper. Three RealSense D435i cameras provide visual observations from base, hand, and third-person viewpoints. The benchmark includes whiteboard wiping and door opening, evaluated with and without a door closer ( 50 N of force to push at the handle).
The primary metric is success rate (\%), where each rollout is evaluated over sequential phases: Approach, Pick/Grasp, Manipulate, Place/Push, and Leave/Cross.

\subsubsection{Baselines.}
To evaluate the effectiveness of our approach in loco-manipulation tasks, we compare FWBC-VLA with representative baselines: OpenVLA~\cite{kim2024openvla}, StarVLA~\cite{Community2026StarVLAAL}, $\pi_{0.5}$~\cite{openpi0.5}, Gr00t N1.6~\cite{grootn1}, ACP~\cite{Hou2024AdaptiveCP},  ForceVLA~\cite{forcevla2025} and FWBC-GT (Using real force data as ground truth).
We use publicly released pretrained models and fully fine-tune them for 50K steps on the WL\&Arm dataset for each task, using the same robot-state observations, camera views, and 16-dimensional output action space.
OpenVLA, StarVLA, $\pi_{0.5}$ and Gr00t N1.6 are foundation models trained on diverse robot data.
We use force from real F/T sensor for ACP, ForceVLA and FWBC-GT, which incorporate force modality into their control policies.

\subsubsection{Tasks.}
To evaluate contact-rich loco-manipulation ability, We evaluate two long-horizon tasks: (i) mobile door opening, with and without a door closer. (ii) Whiteboard wiping.
The ratio of in-domain to out-of-domain stain patterns is 1:2.
Whiteboard wiping is evaluated at the Approach, Pick, Wiping, Place, and Leave stages; door opening is evaluated at the Approach, Grasp, Manipulate, Push, and Cross stages.

\paragraph{Result on contact-rich loco-manipulation (Q3).}
As summarized in Table~\ref{tab:real-world-comparison}, FWBC-VLA achieves the highest
success rate at every reported stage on both tasks. For whiteboard
wiping, it reaches 76\% at Wiping and 64\% final success,
exceeding the strongest stage-specific baselines, ACP and ForceVLA,
by 32 and 40 percentage points, respectively. For door opening,
FWBC-VLA obtains 56\% at Manipulate and 52\% at both Push and
Cross, outperforming ForceVLA by 28 and 40 points. Final-stage
margins reach 40 points on both tasks, showing that the advantage
grows during contact-intensive execution.
Furthermore, comparing FWBC-VLA to FWBC-GT using ground-truth force sensing, the close success rates demonstrate that our estimation is effective and can be used for contact-rich tasks.

Fig.~\ref{fig:force-effect} illustrates baseline failures: contact with the board becomes uneven, and visible ink remains after five minutes of wiping.
The representative rollouts in Fig.~\ref{fig:real-world-rollouts} further show baseline failures following body drift. 
In contrast, FWBC-VLA maintains task progression.
Together, the results support explicit interaction feedback for sustained contact in loco-manipulation.

\subsection{Ablation of Force-Conditioned Compensation}
\label{sec:compensation}

\paragraph{Ablation setup.}
We progressively introduce the additional components built on top of
$\pi_{0.5}$, including the force-aware interface (FI) and bounded body
compensation (BC), and compare their effects across four contact-critical
stages. FI comprises the HSR-Force interaction estimate and force-token
conditioning in the action expert, whereas BC converts the projected
body-load proxy into bounded base corrections. Since BC relies on the
interaction estimate produced by FI, a BC-only configuration without FI
is undefined. We therefore evaluate three valid hierarchical variants.
All variants use the same task data, training budget, initial-condition
distribution, and evaluation protocol.
\paragraph{Component-wise ablations on FI and BC (Q4).}
As shown in Table~\ref{tab:comp_ablation}, adding FI increases the average success rate from 12.0\% to 35.0\%, with clear gains in handle pressing, door pushing without a closer, and board cleaning.
Adding BC further improves the average success rate to 59.5\%, providing the largest incremental gain of 24.5 percentage points. 
Its benefit is concentrated in sustained high-load stages: success increases by 52 percentage points when pushing a door with a closer and by 44 percentage points in board cleaning. In contrast, handle pressing and door pushing without a closer remain nearly unchanged. 
These results demonstrate that FI provides contact awareness for manipulation, while BC is essential for maintaining stable whole-body behavior under sustained external forces.


\section{Conclusion and Future Work}
We present FWBC-VLA, a sensorless force-aware feedback framework for contact-rich loco-manipulation. 
HSR-Force combines historical and current states through a fixed residual gate, summarizes the residual torque as a contact descriptor, and projects the directional residual into EE- and body-frame load proxies.
These signals condition the VLA action expert and compensation sidecar, coupling task-level action with whole-body stabilization. 
Real-world experiments show that explicit interaction feedback and whole-body compensation improve performance relative to the baselines.

\paragraph{Limitations and future work.}
Current frameworks demonstrate the benefits of estimated force, but the impact of force estimation accuracy on task performance warrants further investigation.
Future work will investigate broader contact-aware perception and further develop the tightly integrated whole-body VLA control paradigm for complex loco-manipulation tasks on humanoid and wheeled-base robots.

\bibliography{references}

\end{document}